\documentclass[acmsmall]{acmart}
\acmSubmissionID{3607}

\usepackage{graphicx}
\usepackage{subcaption}
\usepackage{hyperref}
\usepackage{siunitx}

\setcopyright{acmlicensed}
\acmJournal{PACMCGIT}
\acmYear{2023} \acmVolume{6} \acmNumber{2} \acmArticle{} \acmMonth{8} \acmPrice{15.00}\acmDOI{10.1145/3606928}

\begin{document}

\title{Physics-based Motion Retargeting from Sparse Inputs}

\author{Daniele Reda}
\email{dreda@cs.ubc.ca}
\affiliation{
 \institution{University of British Columbia}
 \country{Canada}
}

\author{Jungdam Won}
\affiliation{
 \institution{Seoul National University}
 \country{South Korea}
}

\author{Yuting Ye}
\affiliation{
 \institution{Reality Labs Research, Meta}
 \country{United States of America}
}

\author{Michiel van de Panne}
\affiliation{
 \institution{University of British Columbia}
 \country{Canada}
}

\author{Alexander Winkler}
\email{winklera@meta.com}
\affiliation{
 \institution{Reality Labs Research, Meta}
 \country{United States of America}
}

\renewcommand{\shortauthors}{Reda et al.}

\begin{abstract}
Avatars are important to create interactive and immersive experiences in virtual worlds.
One challenge in animating these characters to mimic a user's motion is that commercial AR/VR products consist only of a headset and controllers, providing very limited sensor data of the user's pose. Another challenge is that an avatar might have a different skeleton structure than a human and the mapping between them is unclear. In this work we address both of these challenges. We introduce a method to retarget motions in real-time from sparse human sensor data to characters of various morphologies. Our method uses reinforcement learning to train a policy to control characters in a physics simulator. We only require human motion capture data for training, without relying on artist-generated animations for each avatar. This allows us to use large motion capture datasets to train general policies that can track unseen users from real and sparse data in real-time. We demonstrate the feasibility of our approach on three characters with different skeleton structure: a dinosaur, a mouse-like creature and a human. We show that the avatar poses often match the user surprisingly well, despite having no sensor information of the lower body available. We discuss and ablate the important components in our framework, specifically the kinematic retargeting step, the imitation, contact and action reward as well as our asymmetric actor-critic observations. We further explore the robustness of our method in a variety of settings including unbalancing, dancing and sports motions.
\end{abstract}


\begin{teaserfigure}
    \centering
    \includegraphics[width=0.9\linewidth]{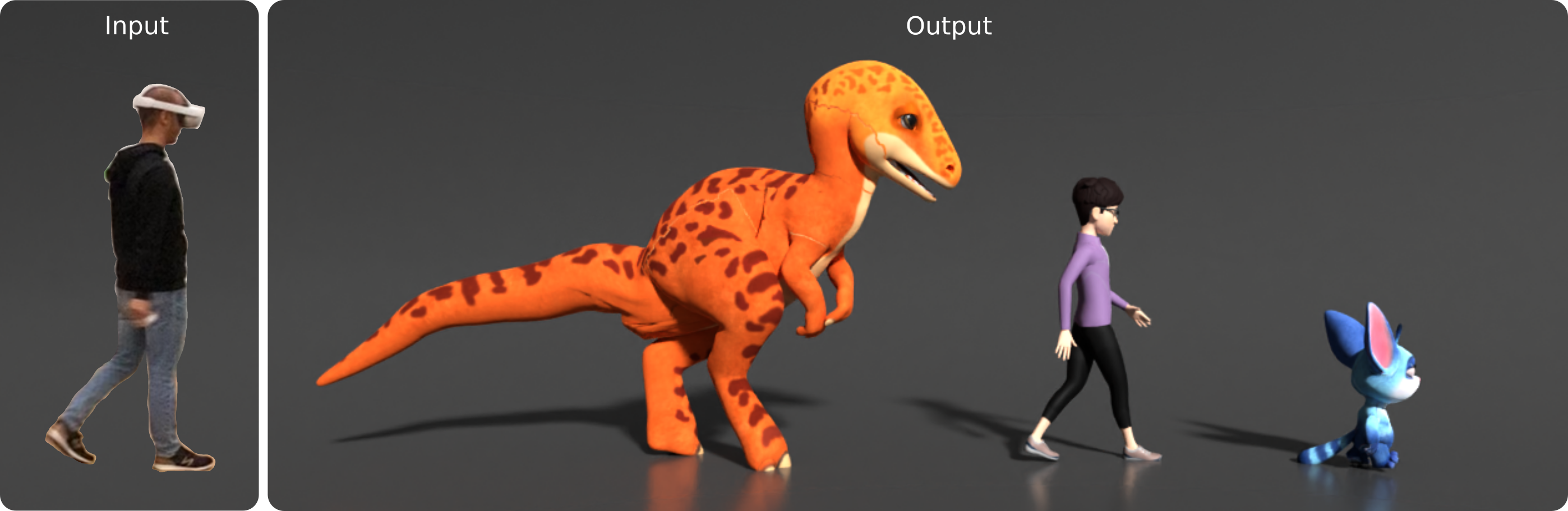}
    \caption{Our method uses only a headset and controller pose as input to generate a physically-valid pose for a variety of characters in real-time.}
    \label{fig:teaser}
\end{teaserfigure}

\maketitle

\section{Introduction}
\label{sec:intro}

Augmented and Virtual Reality (AR/VR) has the potential to provide rich forms of self-expression. By using human characters it is easier to accurately reflect the motions of a user. However, many users might want to portray themselves via non-human characters.
Games with non-human player characters already demonstrate the great appeal of this type of embodiment, 
albeit one that works within the limited immersion afforded by current gaming input devices and displays. 
How can we best allow users to embody themselves in non-human characters using current AR/VR systems?
Our work seeks to make progress on this question. This entails multiple challenges, in particular: 
(a) AR/VR systems provide only sparse information regarding the pose of the user,
  obtained from a head-mounted device (HMD) and two controllers.
(b) The target character may have significantly different dimensions and body types, as shown in \autoref{fig:teaser}; and
(c) Kinematic animation, including that resulting from kinematic retargeting, often lacks physical plausibility, producing movements that lack a feeling of weight. 

We propose a method to address these challenges. In particular, we develop an imitation-based
reinforcement learning (RL) method that uses the sparse sensor input of a user to drive 
a physics-based simulation of the target character. This directly takes into account the physical properties of the given character,
such as the heavy tail of a dinosaur or the short-legs of a mouse character, as shown in \autoref{fig:teaser}.
We only require human motion capture data
for training, without relying on artist-generated animations for each avatar.
This allows us to use large motion capture datasets to train general policies
that can track unseen users from real and sparse data in real-time.
We identify ingredients as being important to successful retargeting in this setting,
including foot contact rewards, sparse mapping of key features for retargeting, 
and suitable reward terms that offer further style control.
Many of the pieces that we rely on exist elsewhere in the literature.
Our primary contribution lies with bringing them together in a way that
enables a new retargeting capability well-suited to current AR/VR systems.
We are the first to show a framework that works with \textit{real} data from \textit{sparse} sensors in \textit{real} time while producing high-quality motions for non-human characters.
We validate our design choices through a variety of ablations.

\section{Related Work}
\label{sec:related}

\begin{figure*}
  \centering
\includegraphics[width=\linewidth]{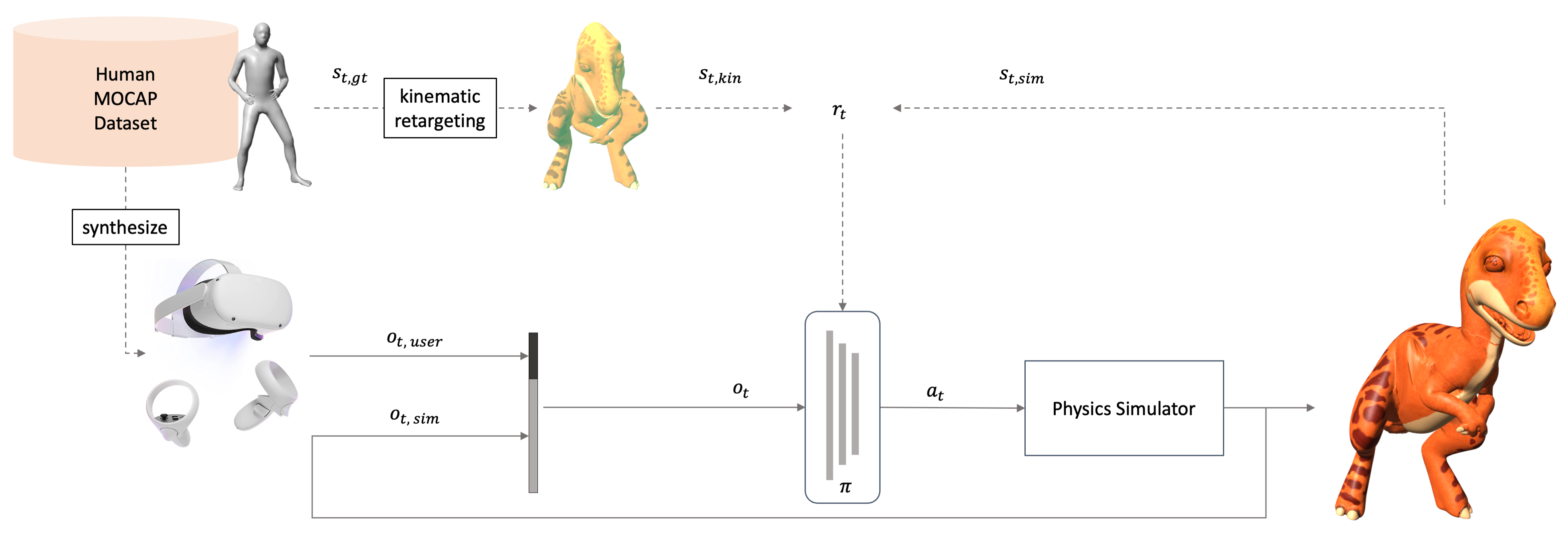}
  \caption{Overview of our system. The policy $\pi$ receives the Quest sensor input $o_{t, user}$ and the current state of the simulated character $o_{t, sim}$ as observation and computes torques $a_t$ to apply to a physics simulator. During training, we use human motion capture data $s_{t, gt}$ to estimate a rough pose $s_{t, kin}$ of the simulated character ("kinematic retargeting"). The reward encourages the simulated character $s_{t, sim}$ to imitate this rough kinematic pose $s_{t, kin}$ as best as possible, while respecting all the physical constraints imposed by the simulator. After the policy is trained, full-body data or kinematic retargeting is not required anymore, and the simulated character can be driven purely by the HMD and controller sparse sensor.}
  \label{fig:system-diagram}
\end{figure*}

In this literature review we focus on the most relevant works in motion tracking, retargeting, and physics-based control.

\subsection{Human Motion Tracking}

Many solutions exist for full-body tracking of human motion, varying in their choice of sensors, 
the number of sensors, and their placement.
Optical marker-based systems with external cameras remain the most common choice for applications requiring high accuracy, e.g.,~\cite{Vicon}.
\textit{Markerless} and \textit{vision-based} approaches rely on cameras alone to generate full body poses. 
Common approaches leverage human body models such as SMPL as a pose prior~\cite{SMPL:SIGA:2015, Kanazawa:2019:CVPR, xu2019denserac, FrankMocap:ICCV:2021},  using extracted keypoints or correspondences from the images~\cite{Densepose:CVPR:2018, OpenPose:PAMI:2019}, or use physics-based priors, e.g.,~\cite{rempe2021humor}
\textit{Wearable} sensors are another common choice, relying on sensors attached on the user's body, such as Inertial Measurement Unit (IMU) devices, e.g.,~\cite{Marcard:SIP:2017, DIP:SIGA:2018, jiang2022transformer}.

When using AR/VR devices, systems are further limited by the sparse sensors available. Most commonly available units are comprised of 3 tracker devices: a head-mounted device (HMD) and two controllers, one for each hand.
As a human motion tracking device, these are handicapped by the lack of sensory information regarding 
the lower body and legs, which are essential to synthesizing believable full-body motion. 
Multiple methods have been proposed to address this, using transformers~\cite{jiang2022transformer, Transfomer:NIPS:2017}, VAEs~\cite{dittadi:ICCV:2021} and normalizing flows generative models~\cite{aliakbarian2022flag}. 
Being kinematic-based approaches, however, these methods do not enforce physical properties and thus 
suffer from motion artifacts such as foot-skating and jitter.
Physics-based approaches have also recently been proposed~\cite{winkler2022questsim,neural3points}. 
These both make use of reinforcement learning and physics to learn general and robust policies that drive full-body 
avatars, conditioned on input from a VR device. These are closest to the work we present in this paper, and
have great promise, although come with their own limitations.
The Neural3Points method~\cite{neural3points} is specific to a single user and 
uses auxiliary losses and an intermediate full-body pose predictor. 
Relatedly, \citet{winkler2022questsim} proposes a more direct approach that is able to control a simulated
human avatar and generalizes to users of different heights and multiple type of motions. 
Our work generalizes the method of~\citet{winkler2022questsim} in two important ways: (1) we learn \
physics-based retargeting to characters having different morphologies, and 
(2) we enable real-time retargeting. 

\subsection{Retargeting Motions}

The motion retargeting problem is that of remapping motion from a source character or skeleton,
often driven by motion capture data, to another character of possibly different dimensions.
This is a long-standing problem for which many solutions have been proposed.
Arguably the most challenging version of this problem arises when the source and target characters may
differ significantly in terms of their morphology and skeleton, as is also the case for our work.

Kinematic retargeting methods often approach the problem by allowing the user to specify directly, or alternatively to learn via examples, 
a model for source-to-target pose correpondences, e.g., \cite{monzani2000using,yamane2010animating,seol2013creature}.
This creates a puppetry system, where target motions can be further cleaned to respect contacts with the help
of inverse kinematics.  Kinematic motion deformation approaches can be used to adapt multiple characters trajectories
for motions involving coordination such as moving boxes~\cite{kim2021interactive}.
Recent work proposes a kinematic method to learn how to retarget without requiring any explicit pairing between motions~\cite{aberman2020skeleton},
and this is also demonstrated to work on skeletons with very different proportions.
Other recent work examines how to learn efficient kinematic motion retargeting for human-like skeletons
while preserving contact constraints, such as when hands and arms have self-contact with the body~\cite{villegas2021contact}.

Physics-based retargeting methods aim to produce a physics-based simulation of the output motion,
which results in crisp contacts and physically-plausible motion of the target character.
An offline approach to motion retargeting using spacetime trajectory optimization is presented in~\citet{al2018robust}.
The final output uses LQR trees, and thus the given motions can cope with some perturbations.
A method is recently proposed for using interactive human motion to drive the motion of a quadruped robot~\cite{kim2022human}.
A curated dataset of matching pairs of human-and-robot motions is used to develop relevant kinematic mappings
for particular motions or tasks. A deep-RL policy is then learned that can track the target kinematic motions in real time,
enabling a form of real-time human-to-real-robot puppetry.  In our setting, we assume significantly sparser user input and motion specifications.

\subsection{Physics-based Character Simulation}

Controllers for physics-based characters have been extensively explored.
The ability to imitate reference motions was first demonstrated to varying extents
in a number of papers over the past 15 years, e.g., \cite{yin2007simbicon,lee2010data,ye2010optimal,coros2010generalized,liu2010sampling,geijtenbeek2012simple}.
These methods often incorporated some iterative optimization to adapt to a specific motion
and used a simple control law to provide robust balance feedback.
Some of these methods were also adapted to produce motions for non-human characters, 
e.g.,~\cite{geijtenbeek2013flexible,wampler2014generalizing}.


Neural network policies, trained via deep reinforcement learning (RL), provide new capabilities
to learn new skills from scratch, or to imitate artist-provided motions or motion capture clips, 
e.g., \cite{2017-siggraph-deeploco,won2017train,2018-siggraph-deepmimic}, including demonstrations for non-human characters.
More recent methods provide more flexibility in sequencing motions for basketball~\cite{2019-SiggraphAsia-basketball}
or, more generally, to track online streams of motion capture data~\cite{chentanez2018physics,2019-siggraph-drecon,won2020scalable,2021-siggraphasia-supertrack}.
Control policies have also been learned which are conditioned on not only the desired motion, but also
the specific morphology of a simulated character, which can then even be changed at run time~\cite{won2019learning}.
We further refer the reader to a recent survey of RL-related animation methods~\cite{kwiatkowski2022survey}.
We build on the foundations provided above for our specific problem,
namely how to retarget from sparse (and therefore potentially highly ambiguous) input data to a 
non-human physics-based character with very different dimensions and proportions.




\section{Method}
\label{sec:method}

An overview of our system is shown in \autoref{fig:system-diagram}. We use reinforcement learning to learn a policy that generates torques for a physics simulator.
During training, we use human motion capture data to both synthesize HMD and controllers data for the policy, and to build a reward training signal. In the following we give an overview of reinforcement learning and then describe each component in detail.

\subsection{Reinforcement Learning}
We use deep reinforcement learning (RL) to learn a retargeting policy for each character.
In RL, at each time step $t$, the control policy reacts to an environment state $s_t$ by performing an action $a_t$. Based on the action performed, the policy receives a reward signal $r_t = r(s_t, a_t)$. In deep RL, the control policy $\pi_\theta(a | s)$ is a neural network. The goal of deep RL is to find the network parameters $\theta$ which maximize the expected return defined as follows:
\begin{align}
J_{RL}(\theta) = \mathop{\mathbb{E}} \left[\sum_{t=0}^{\infty}\gamma^t{r({s}_t, {a}_t)} \right],
\end{align}
where $\gamma \in [0, 1)$ is the discount factor. Tuning $\gamma$ affects the importance we give to future states.
We solve this optimization problem using the proximal policy optimization (PPO) algorithm~\cite{2017-PPO-Schulman}, a policy gradient actor-critic algorithm.
A review of PPO algorithm is provided in \autoref{appendix:ppo-summary}.

\subsection{Characters}

\begin{figure}
  \centering
  \includegraphics[width=0.85\linewidth]{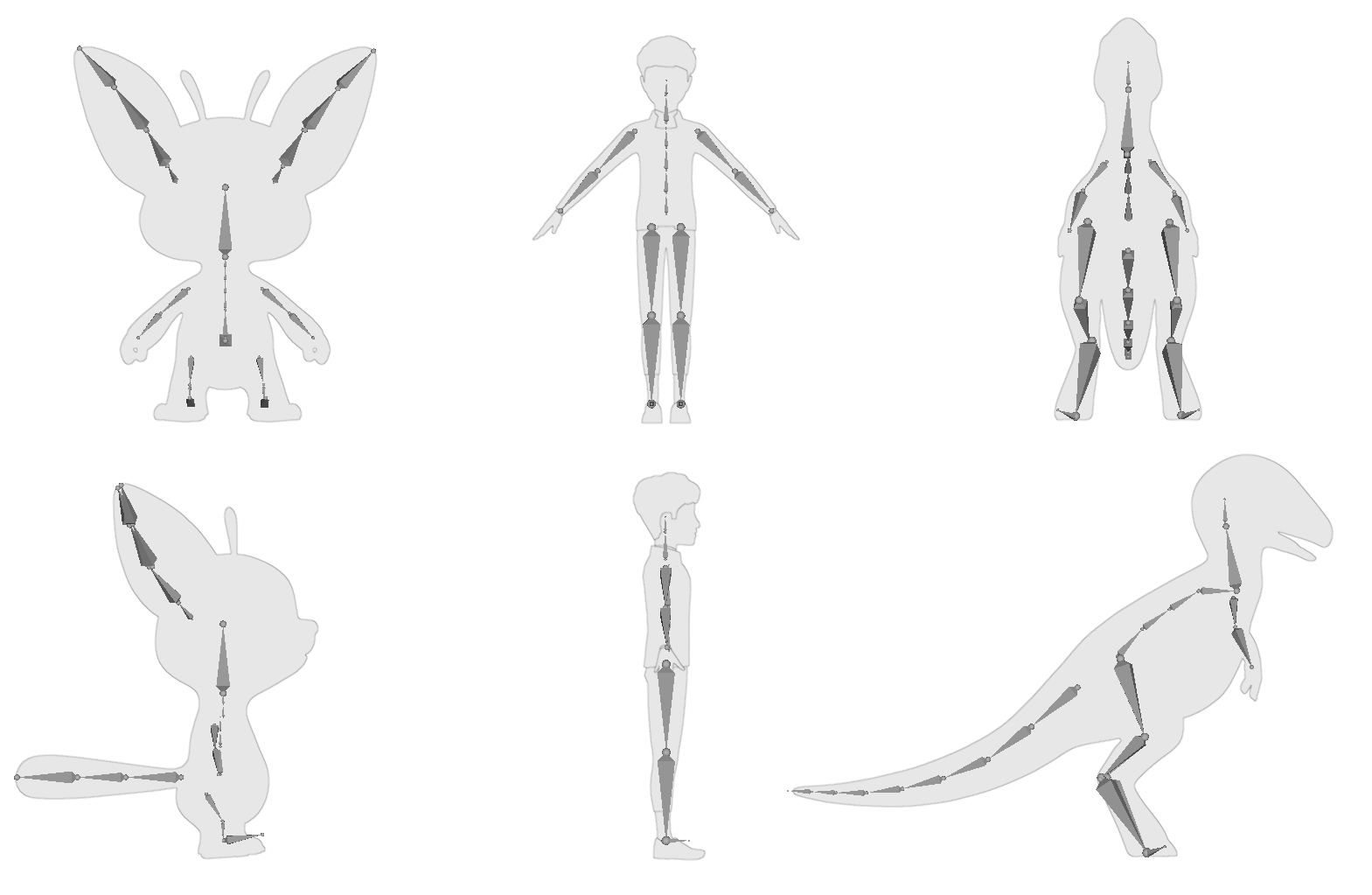}
  \caption{We demonstrate our retargeting solution on three different characters (from left to right): a mouse-like creature named Oppy, a human named Jesse, and a dinosaur we call Dino.}
  \label{fig:characters}
\end{figure}

We demonstrate our retargeting solution on three characters with unique features:
Oppy~\cite{oppy_repo} is a mouse with a short lower body, a big head, big ears and a tail;
Dino is a tall dinosaur, with a long and heavy tail and head, and short arms;
Jesse is a human-like cartoon character with a skeleton structure similar to the mocap data.
\autoref{fig:characters} shows a visual representation of the characters and \autoref{table:characters} details the structure of their skeletons.

\begin{table}[H]
\centering
\small
\caption{Character details.}
\label{table:characters}
\begin{tabular}{cccc}
\toprule
\bf{Parameter}           & \bf{Oppy} & \bf{Dino} & \bf{Jesse} \\
\midrule
Weight (kg)              &  7     &  180      &  60           \\
Height (cm)              &  80    &  250      &  180          \\
Total links              &  24    &  30       &  16           \\
Total DOF                &  58    &  44       &  32           \\

Upper body joints        &  7     &  10       &  10           \\
Lower body joints        &  6     &  8        &  6            \\
Tail joints              &  3     &  8        &  -            \\
Ear joints (x2)          &  4     &  -        &  -            \\
Max Torque               &  40    &  300      &  300          \\
\bottomrule
\end{tabular}
\end{table}




\subsection{Observations}
The observation contains two parts: simulated character data $o_{t, sim}$ and user's sparse sensor data $o_{t, user}$.
\begin{align}
    o_t &= [o_{t, sim}, o_{t-1, user}, o_{t, user}]\\
    o_{t, sim} &= [o_{sim, q}, o_{sim, \dot{q}}, o_{sim, x}, o_{sim, R}]\\
    o_{t, user} &= [h_t, l_t, r_t, R_{h, t}, R_{l, t}, R_{r, t}]
\end{align}
The simulated character's state is fully observable in the simulation. Therefore, even though the sensor signals is sparse, the policy can still rely on the full state of the simulated character.
This observation consists of joint angles $o_{sim, q} \in \mathbb{R}^j$ and joint angle velocities $o_{sim, \dot{q}} \in \mathbb{R}^j$ of all degrees of freedom $j$ of the character.
We also provide Cartesian positions $o_{sim, x} \in \mathbb{R}^{l\times3}$ and orientations $o_{sim, R} \in \mathbb{R}^{l\times6}$ of a subset $l$ of links of the character.
The orientations consist of the first two columns of their rotation matrices. 
All positions and orientations are expressed with respect to a coordinate frame located on the floor below the character which rotates according to the character heading direction. This is useful to make the controller agnostic to the heading direction.

The sensor data, either coming from the real device or synthetically generated from the training data (described in \autoref{sec:synthetic-data}), consists of the position and orientation of the HMD $h$, the left controller $l$ and the right controller $r$. Positions and orientations are expressed in the same coordinate system as the simulated character observations. To allow the policy to infer velocities, we provide it two consecutive sensor observations $[o_{t-1, user}, o_{t, user}]$.

Inspired by~\citet{pinto2018asymmetric}, we use asymmetric observations.
At training time we augment the value function observation by providing the full human mocap pose and future human mocap state information.
This complete view of the state allows the value function to better estimate the returns. The better the return estimate, the easier it is for the policy to learn. We are allowed to provide this mocap state information, because the value function is required only for training. Real-time inference still only relies on the policy, which uses the sparse sensor input. We ablate this in \autoref{sec:value-ablation} and find that is essential for sparse real time retargeting.

\subsection{Synthetic Training Data}
\label{sec:synthetic-data}
During training, we require HMD and controller data for the observation paired with kinematic poses for each character $s_{t, kin}$ from which the reward $r_t$ is computed.
To synthetically generate the HMD and controller data we offset the mocap head and wrist joints to emulate the position and orientation of HMD, left and right controllers as if the subjects were equipped with an AR/VR device. 

\begin{figure}
  \centering
  \includegraphics[width=0.99\linewidth]{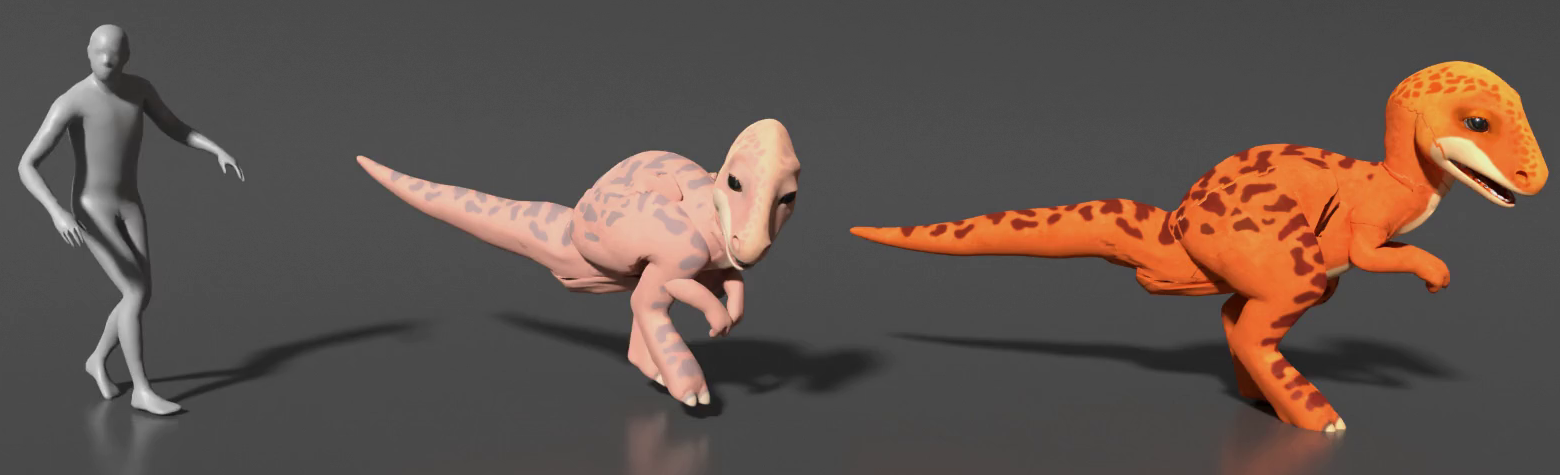}
  \caption{Training data is generated through kinematic retargeting. The left character is the human mocap data. The middle character shows a rough kinematic retargeting by matching selected joint angles. This pose has many artifacts, such as feet sliding due to different leg lengths, self-collisions, floor collisions, and no motion of the tail and ears. The right character is the closest simulated pose that also respects all physical constraints. Notice how the head does not perfectly follow the human, as it's heavier and takes more time to react, no having access to future information but only past and present.}
  \label{fig:tennis-dino}
\end{figure}

Importantly, our system does not require artist-generated animations for each specific character as training data, which would be infeasible to create with the diversity and quantity we require.
Instead we reuse existing human motion capture data $s_{gt}$ and perform a rough kinematic retargeting $s_{kin}$ to the morphology of the simulated character (\autoref{fig:tennis-dino}).
In this step, we manually match selected joint angles of the human to conceptually similar joints of the creature.
For joints where no correspondence can be found, we just set them to their default pose (e.g. ears and tails).
This provides a rough estimate of the creature's motion. However, this motion has many artifacts, such a feet sliding due to different leg lengths, self-collisions, floor collisions, and no motion of the tail and ears.
Nonetheless, we can still use it as a reward signal to train our simulated character.
The physical constraints imposed by the simulation then remove remaining artifacts. Importantly, after the simulated character is trained, it is driven only by a headset and controllers, without requiring any full-body information of the user or any kinematic retargeting.

\subsection{Reward}
\label{sec:reward}
The goal for the simulated character is to imitate the human motion as closely as possible, while respecting all the constraints imposed by physics. Our reward function includes a component for imitation, contact, and action regularization:
\begin{align}
    r_t &= r_t(\text{imitation}) + r_t(\text{contact}) + r_t(\text{action}) \label{eq:reward-general} \\
    r_t(\text{imitation}) &= r_t(q) + r_t(\dot{q}) + r_t(x) + r_t(\dot{x}) + r_t(\text{orientation}) \label{eq:reward-imitation} \\
    r_t(\text{action}) &= r_t(\text{action diff}) + r_t(\text{action min}). \label{eq:reward-action}
\end{align}
Each of the reward terms is expressed using a weighted Gaussian kernel:
\begin{equation}
    r_t(s) = w_s \mathrm{e}^{-k_s d(s_{t, sim}, s_{t, kin})} \label{eq:reward-component}
\end{equation}
where for each term only the specific component of the state $s$ is considered and $d(s_{sim}, s_{kin})$ represent the distance metric between the simulated and kinematic components of the state, $k$ is the sensitivity of the Gaussian kernel, and $w$ is the weight of the reward component. Parameter values and details of the distance metrics for each term are provided in \autoref{appendix:reward}.

\subsubsection{Imitation Reward}
This reward matches the available information between the simulated character $s_{sim}$ and the kinematically retargeted ground truth pose $s_{kin}$.
The five terms represent a weighted sum of the difference between the matching joint angles ($q$), joint angle velocities ($\dot{q}$), Cartesian coordinate positions ($x$) and velocities ($\dot{x}$), and orientation.
The imitation reward term captures the degrees of supervision we want to transfer between human motion data and the simulated character.
For clarity, \autoref{eq:reward-imitation} is the general form which includes all possible terms, but the way they are used differs according to each character.
The less supervision the imitation term provides, the more we rely on physics and the other components to generate a sensible motion. 

Depending on the quality of our kinematically retargeted pose, we can choose which of the aspects of the pose we want the simulated character to imitate more closely.
The least amount of supervision consists in only tracking their root position, which according to our experiments does not produce high-quality motions.
On the other extreme, we also do not want to track every aspect of the kinematically retargeted pose.
For example there is no tail motion in the human mocap data, so the kinematically retargeted pose has all tails set to a stiff default pose. However, a simulated character might want to move the tail to achieve balance and smoother motion. So we do not require these parts of the skeleton to imitate the kinematic pose.

Orientations are skeleton independent, so we rely on the actual human mocap data, not the kinematically retargeted pose to formulate the orientation rewards.
We always formulate a reward that matches the characters root with the human mocap root, as well as the characters head orientation with the human head orientation.
Ablations without these terms are provided in \autoref{sec:reward-ablation}.


\subsubsection{Contact Reward}
The contact reward is a boolean value that checks whether the simulated character's foot contact and the human's foot contact coincide.
We estimate contact of the mocap data based on a velocity and height threshold. For the simulated character, we can directly access contact forces from the simulator and threshold those. 
In most cases the kinematically retargeted leg motion has a variety of artifacts, such as feet sliding or penetrating the ground. Imitating this pose is not physically-valid. Since this reward doesn't depend on the skeleton structure, it can be used for all bipedal characters equally and directly computed from human mocap.
The contact reward is important to give further training supervision and generate the high-quality motions shown. Ablations are provided.  

\subsubsection{Action Reward}
The action reward is a regularization term to minimize total amount of energy consumed by the character.
It consists of two terms that minimize the difference in torque between two subsequent actions and minimize the absolute action value and is defined as:
\begin{gather}
    r_t(\text{action diff}) = \frac{1}{N}\sum_i^N (a_{t-1, i} - a_{t, i})^2 \\
    r_t(\text{action min}) = \frac{1}{N}\sum_i^N a_{t, i}^2 
\end{gather}
where $N$ is the total number of action values which the policy outputs.
The purpose of these components is to incentivize overall lower energy movements and to minimize twitching with a smoother movement between poses.

\subsection{Termination}
\label{sec:termination}
As noted in multiple previous works~\cite{2018-TOG-deepMimic, 2020-envdesign}, early termination techniques are important for learning complex motions through reinforcement learning.
We reset the environment when one of the following two termination conditions is satisfied: the character enters an unrecoverable state, which we define as falling and touching the ground with the upper body, or when the character root position is more than 30cm apart from the scaled root of the motion capture data.
Furthermore, to mitigate the imbalance of visiting and learning to retarget only the early parts of the motion trajectories, we reset the character every 500 steps. We randomly sample a pose from the human data and set the character using the kinematically retargeted pose.

\subsection{Learning Control Policies}

The policy for each simulated character outputs torque values in the range $[-1, 1]$ which are then rescaled according to minimum and maximum torque values for each joint (provided in \autoref{appendix:torque-limits}).
We find this to perform better and be more clear with respect to outputting PD target angles, as shown by previous works~\cite{2020-envdesign}.
We train the policy with PPO and PyTorch auto differentiation software~\cite{2017-PPO-Schulman, paszke:2019:pytorch} and simulate physics with NVIDIA PhysX Isaac Gym physics simulator~\cite{2021-IsaacGym}.
A complete set of hyperparameter details for reproducibility are summarized in~\autoref{appendix:hyperparameters}.

\section{Results}
\label{sec:results}

\begin{figure}
  \centering
  \includegraphics[width=0.75\linewidth]{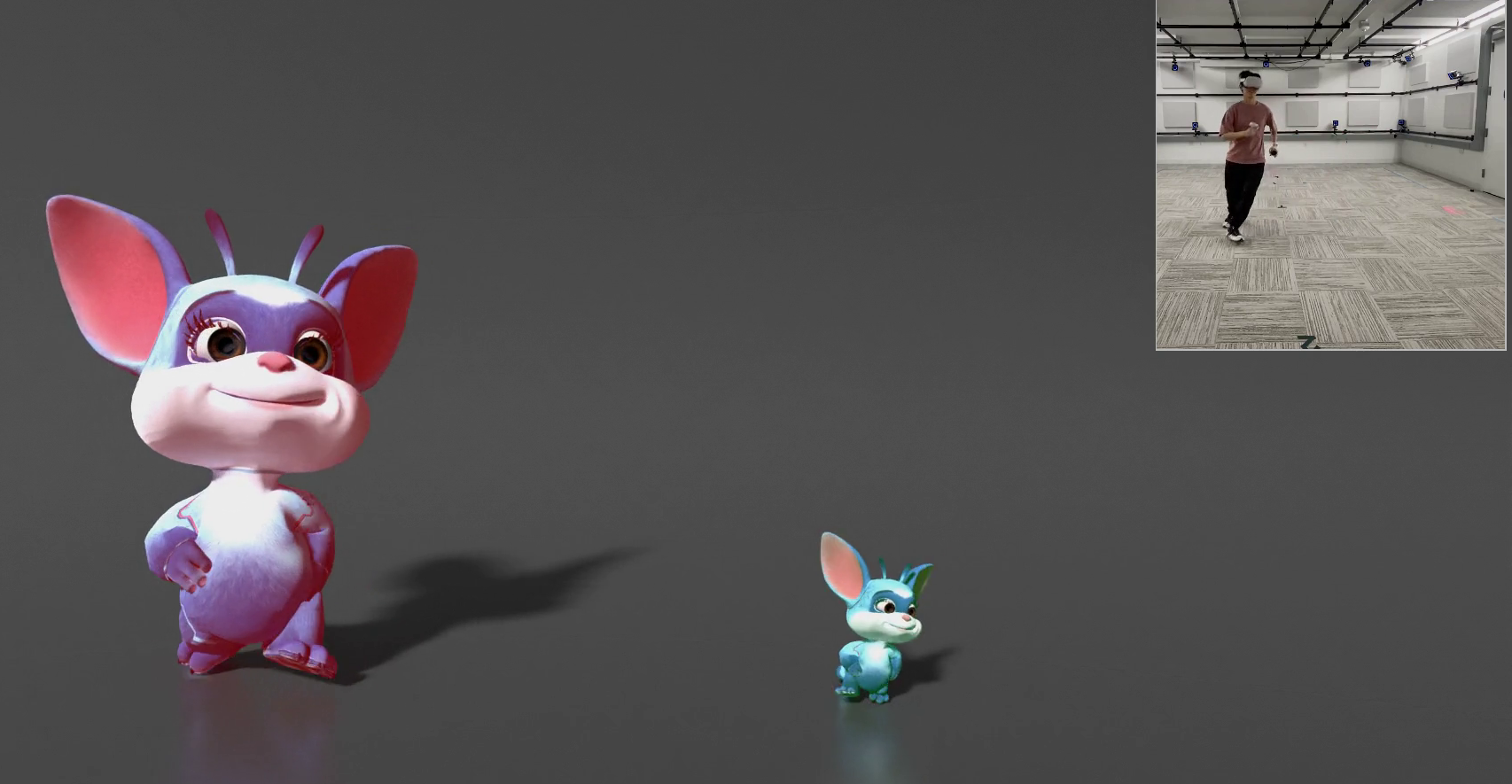}  
  \caption{If the character size matches the user, joint angles and foot contacts between the character and the user are more similar (left). 
  If the simulated character has very different morphology (e.g. here much smaller), the kinematic-retarged pose is less accurate and is mostly ignored by the simulated character in order to generate a physically valid motion. Here the character has to take many steps for a single human step to match the root translation.}
  \label{fig:contact}
\end{figure}

\begin{figure}
  \centering
  \includegraphics[width=0.75\linewidth]{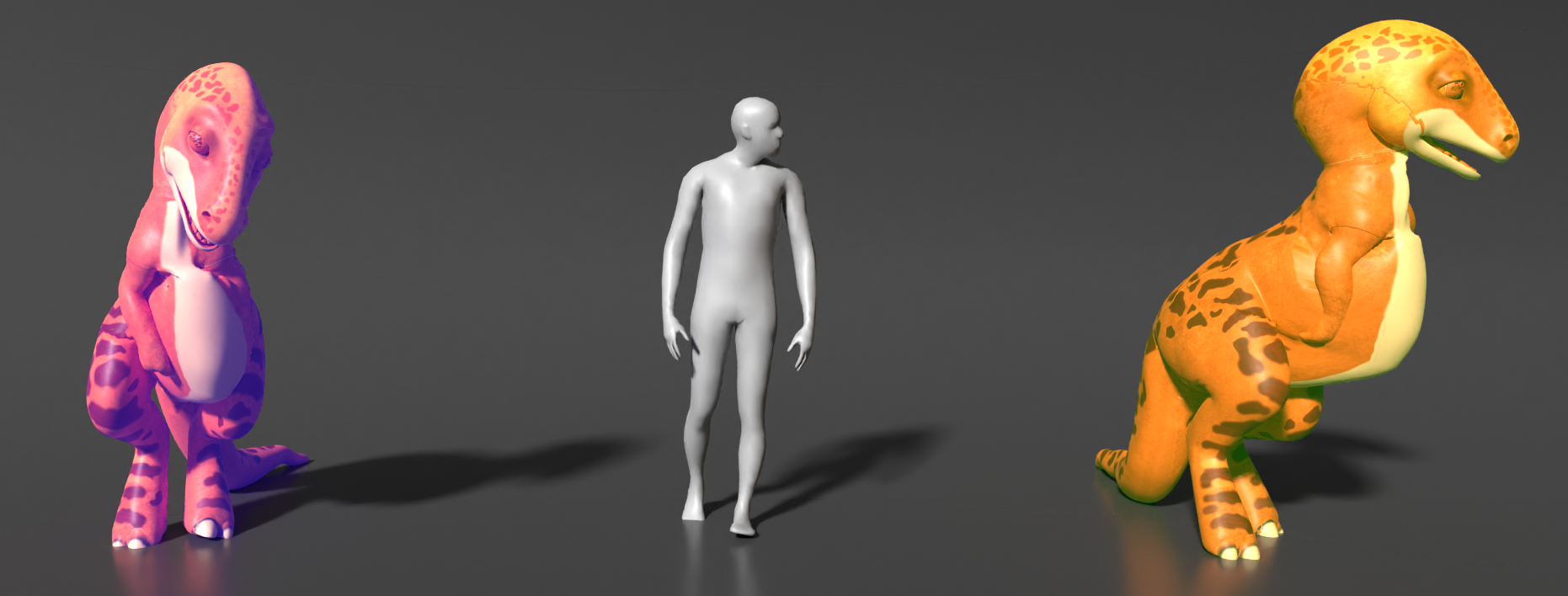}
  \caption{Right Dino has the orientation reward, while Left Dino does not. As the user turns its head, Right Dino follows more closely.}
  \label{fig:orientation}
\end{figure}

\begin{figure*}
    \centering
    \includegraphics[width=\linewidth]{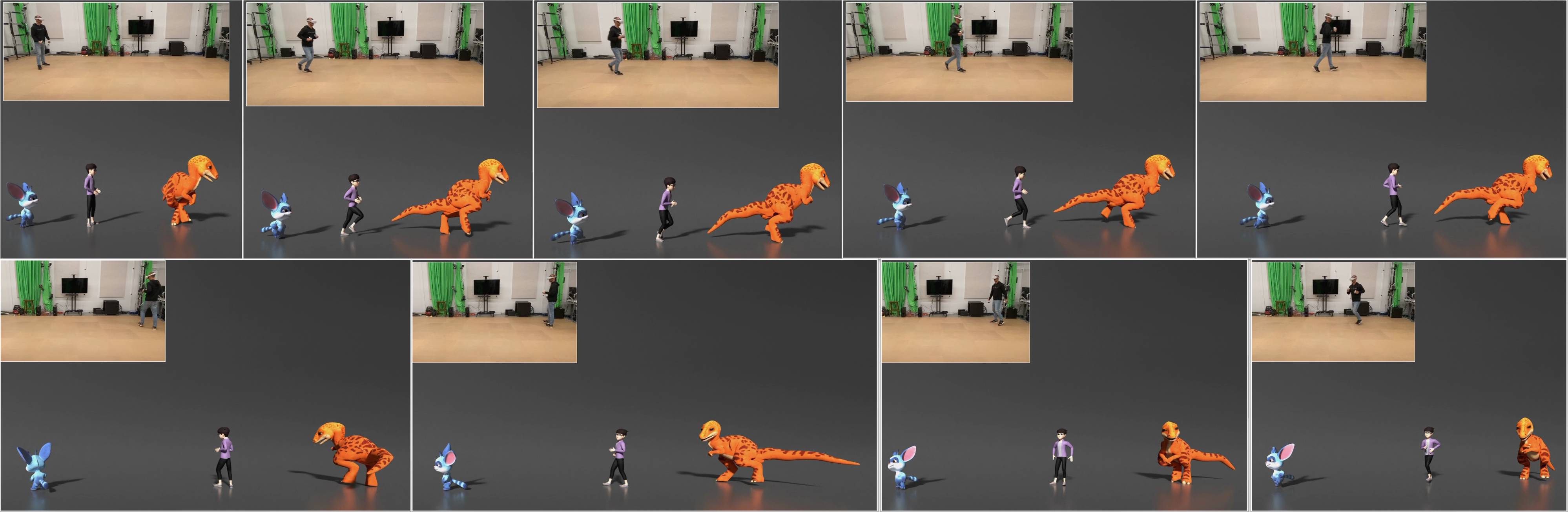}
    \caption{Sequence of frames showing all three characters being controlled in real time with sparse sensory input. Lower-body motion perfectly matches the one of the user and feet contacts are correctly estimated. Watch the accompanying video for more results.}.
    \label{fig:result-sequence-realquest}
\end{figure*}

All experiments are performed on a single 12-core machine with one NVIDIA RTX 2070 GPU.
All models are trained for 24 hours which translates to approximately 6 billion environment steps.

We demonstrate comparable results with two different motion capture datasets.
Our in-house mocap data consists of 4 hours of motion clips of 120 subjects. Specifically, the dataset contains 130 minutes of walking and 110 minutes of jogging.
We also demonstrate robust and general results with the Ubisoft La Forge Animation (LaFAN1) dataset~\cite{2020-lafan}, an open source motion capture dataset containing 5 subjects and 77 sequences. For the purposes of this work, we only considered actions themed \textit{Walk} and \textit{Run}, which consist of a total of 15 sequences and 74 minutes of data.
We note that these motions are very different from the ones in our in-house dataset, containing diverse and hard behaviors and gaiting styles.
At inference, we provide input to the policy with a Meta Quest headset and controllers device.

\subsection{Real-time Retargeting with Headset and Controller}
We thank the QuestSim~\cite{winkler2022questsim} authors for providing us with testing data and video references.
With our method, we are able to control different characters in real time with only headset and controller information.
Importantly, we are able to estimate the lower-body pose of the user from only three points in the upper body and correctly match the user action while transferring it to a character with a different morphology.
Our virtual characters respect physical behaviors and do not suffer from jittering, foot sliding or penetration.
Moreover, we are able to generalize to users not present in the training set and users of different heights.
In \autoref{fig:result-sequence-realquest} we show a sequence in which all three characters are controlled by an unseen user.

\subsection{Retargeting using only Headset}
Some VR systems provide only a head-mounted device (HMD), without the two controllers. This provides an even more challenging domain, requiring the policy to predict a full-body pose and control a virtual character from a one-point input.
Nonetheless, our trained models are robust to the lack of controller signal and are able to retarget real time user data from unseen users even from this extremely sparse input, albeit with a lower quality compared to before. We invite the reader to watch the video available in the supplementary material.

\subsection{Reward component ablations}
\label{sec:reward-ablation}
Some reward components are essential to get good motions. Here we go through a few interesting examples.

\subsubsection{Contact Reward}

The contact reward shapes the gait style of the character. Both Oppy and Dino display different locomotion behaviors when using this reward component. Furthermore, as the character size changes, more signal can be transferred to the simulated character. In \autoref{fig:contact} we show Oppy in two different sizes. When Oppy's size matches the user, it performs the same gait style and distance motions; when it is smaller, by matching the correct gait style it will travel less distance, while it can perform a faster gait to keep up with the user, depending on the weighting of the reward components.
Similarly, in \autoref{fig:result-sequence-realquest}, the different frames show the matching gaits between the three characters and the user.

\subsubsection{Orientation Reward}

Providing signal for mimicking head and root orientation is an essential component to support more fidelity in tracking user's head and overall movements. We show in \autoref{fig:orientation} how Dino without the head orientation component is unable to correctly move its head in the same way as the user. As shown in the supplementary video, both Oppy and Dino without head orientation reward component show the head wobbling left and right while walking. These characters have heavy heads needing learned control.
\section{Discussion}
\label{sec:discussion}

We discuss different capabilities and components of our system.

\subsection{Physics-based control}
\begin{figure}
  \centering
  \includegraphics[width=0.75\linewidth]{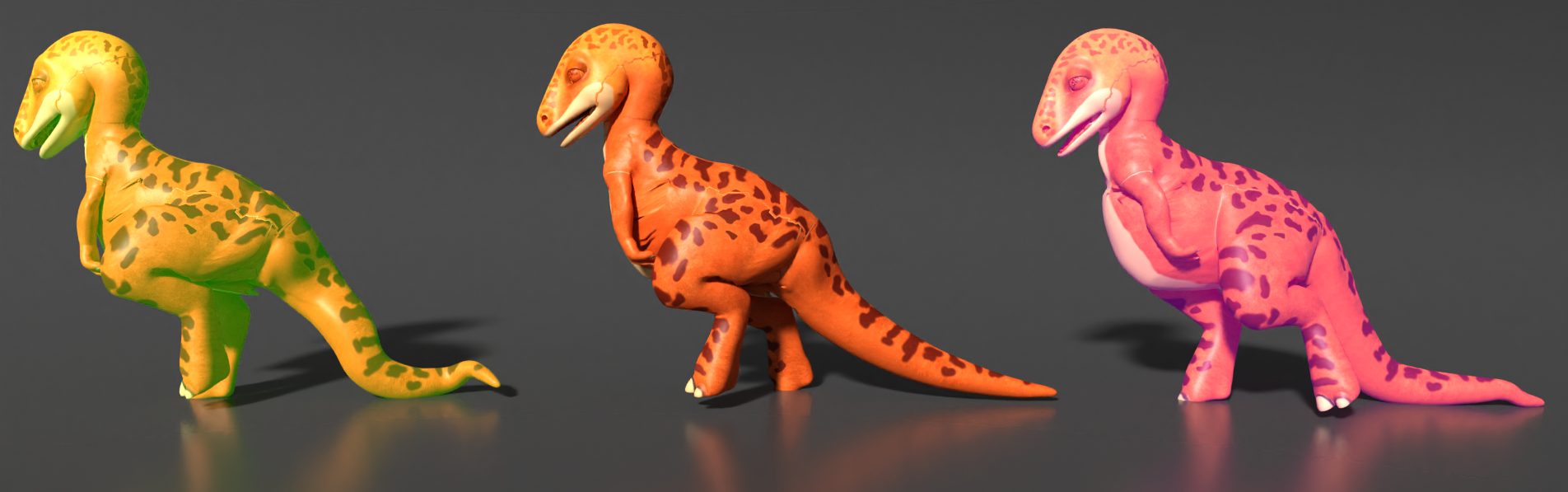}
  \caption{Left Dino's tail has 2 active joints and the remaining 6 passive; Center Dino's tail has 2 active joints and the remaining 6 fixed; Right Dino's tail is completely passive.}
  \label{fig:tail}
\end{figure}
Physics acts as a powerful helper in driving the motion of components with missing pose information, with the skeleton description as underlying prior.
For the tail of Dino, the simulator affords several stylization options, i.e., whether we allow more joint mobility and passively actuate it through a PD controller with fixed-set input as secondary motion or we let the policy make active control decisions.
In \autoref{fig:tail} we show three examples, in which Dino's tail is fixed, passively actuated, or controlled by the learned policy.
Tail and ears of Oppy are all treated as secondary dynamics.
This stylization would not be possible in a kinematic retargeting setting.

\subsection{Controlling the style}
Our method is robust to different set of parameters. Once changed, most parameters still output a reasonable motion controller with different styles. As described in \autoref{sec:reward-ablation}, the contact reward shapes the gait style of the character, and modified together with the size of the character would produce different gait styles.

The kinematic retargeting described in \autoref{sec:synthetic-data} only requires a rough retargeting to produce sensible motions, as the physics dynamics correct the artifacts. Moreover, tuning the key joints for the kinematically retargeted motion produces an overall modification of the style. For example, it is possible to give Dino a more horizontal feeling, with the tail straight behind the back and not touching the ground, by tuning the spine parameter to be more bent over. An illustration of this tail is provided in \autoref{fig:tennis-dino} and in the supplementary video, and noticeable difference can be observed compared to \autoref{fig:tail}.

\subsection{Importance of asymmetric observations}
\label{sec:value-ablation}
During training we provide a richer observation to the value function compared to the one we provide to the policy. Specifically, while at inference the controller receives only real time sparse information (i.e. no future and no full-body pose), there is no need to constrain the value function since at training time this signal is available.
In our experiments, we notice that the outcome of training a policy with a value function that receives no future and no full body pose, is an overall less robust policy. It is able to retarget easy walking examples coming from the training data, but it fails at harder motions like running and is incapable of generalizing to real data coming from an unseen user.

\subsection{Quality of open-source datasets}
We test our method with two different datasets, a 4 hour in-house dataset and a 74 minutes open-source dataset.
While we notice that a larger and more diverse dataset improves the quality of the final motions, models trained with either of these datasets are robust and capable. Both are able to generalize to unseen users, and perform in real-time, even with headset-only sensory input.

\section{Conclusions}

We have presented a method to retarget a user's motion to simulated characters,
in a challenging setting:  the target characters can differ significantly in size and body morphology;
we require a real-time remapping; and the mapping needs to be driven
by the sparse motion data coming from an AR/VR device.
We show that physics-based simulations, driven by asymmetric actor-critic RL policies, 
allow for effective retargeting in this difficult setting.
The motions generated by the policies track those of the user while also 
being appropriate to the physics of the target character.
We introduce a general reward description which allows for tuning of the degree of supervision 
and adapts to a range of character morphologies.
Numerous ablations allow us to understand the impact of various parameters and design choices,
including varying degrees of available tracking information, the impact of contact rewards,
choices related to the secondary motion of tails and ears, and more.

Our work still has a number of limitations.
Our controller fails to track challenging motion sequences, where the user performs fast and dynamic movements or uncorrelated upper/lower body motions.
In these scenarios, a kinematic-based controller acting directly in the pose space will still be able to produce a motion, 
albeit not of high quality, and it will be able to catch up as parts of the motion become easier by "teleport" between poses without correlation.
Instead, our controller has to produce a correct sequence of joint torques to control the character and may suffer from compounding tracking errors until it fails.
An approach that divides the pipeline in two stages, similar to~\citet{neural3points} where first a network predicts the full-body pose and then a high-frequency controller outputs torques, could allow to regain the advantages of kinematic-based systems 
when needed.
While our framework allows richer forms of self expressions for users, empowering them to control different kind of characters, we are only scratching the surface of the complexities that arise due to different target skeletons. 
Our characters are still bipeds.
Increased character complexity might be achieved by supplying skeleton information to the policy~\cite{won2019learning}, using graph neural networks to learn a flexible policy similarly to~\citet{wang2018nervenet}, or training an auxiliary network to find a mapping between source and target skeletons.



\bibliographystyle{ACM-Reference-Format}
\clearpage
\bibliography{bibliography}

\clearpage
\appendix
\section{Reward Details}
\label{appendix:reward}

Parameter values for each term of \autoref{eq:reward-general} and \autoref{eq:reward-component} are provided in \autoref{appendix:reward}.

\begin{table}[H]
\centering
\small
\caption{Reward parameters for each character.}
\begin{tabular}{cccc}
\toprule
\bf{Parameter}          & \bf{Oppy} & \bf{Dino} & \bf{Jesse} \\
\midrule
$w_{q}$                  &  1     &  1        &  4           \\
$k_{q}$                  &  20    &  20       &  25          \\
$w_{\dot{q}}$            &  0     &  0        &  0.5         \\
$k_{\dot{q}}$            &  -     &  -        &  1           \\
$w_{x}$                  &  1     &  1        &  2.5         \\
$k_{x}$                  &  6     &  6        &  6           \\
$w_{\dot{x}}$            &  0     &  0        &  0.7         \\
$k_{\dot{x}}$            &  -     &  -        &  2           \\
$w_{\text{contact}}$     &  1.5   &  1.5      &  0           \\
$k_{\text{contact}}$     &  1     &  1        &  -           \\
$w_{\text{orientation}}$ &  1     &  1        &  2           \\
$k_{\text{orientation}}$ &  3     &  3        &  3           \\
$w_{\text{action diff}}$ &  2     &  2        &  1.5         \\
$k_{\text{action diff}}$ &  150   &  150      &  10          \\
$w_{\text{action min}}$  &   0.2  &   0.2     &  0.5         \\
$k_{\text{action min}}$  &   25   &   25      &  25          \\
fail reward              &   -5   &   -5      &  -5          \\
\bottomrule
\end{tabular}
\end{table}

Given the state of the simulated character and the ground truth pose coming from the motion capture dataset, the distance metric for the different imitation reward components is formulated as a weighted sum of the Euclidean distance between the two values:
\begin{equation}
    d(x_{sim}, x_{gt}) = \sum_i w_i \lVert q_{x, sim} - q_{x, gt} \rVert_2^2
\end{equation}

where $i$ represent the joint angles or the link positions and weights vary according to the character.
As described in \autoref{sec:reward}, the imitation reward defines the degree of supervision.
As the two characters are closer alike, we can rely more on this reward. For Jesse, in fact, all joint weights are equal to 1. For Oppy and Dino, which have a different lower body size compared to a human, we rely more on the style reward for a good motion and decrease the weight of all lower body joints to 0.3.
For link weights, for Oppy and Dino we set all weights to zero other than for the root, which is set to 1, for Jesse we track also end effectors.

Contact distance metric is also computed through the Euclidean distance between ground truth human motion data and simulated character data.
We define that a human foot is in contact if its height is less than 20cm above the ground and the norm of its velocity is less than 0.4 m/s.
For the simulated character, a force threshold of 1 N is set on the feet link.

The orientation distance metric, given the two orientations in quaternions, first computes the composition of the ground truth quaternion with the inverse of the simulated quaternion. Then, takes the distance norm of its axis angle representation. 

\section{Proximal Policy Optimization}
\label{appendix:ppo-summary}
Let an experience tuple be $e_t = (o_t, a_t, o_{t+1}, r_t)$ and a trajectory be $\tau = \{e_0, e_1, \dots, e_T\}$. 
We episodically collect trajectories for a fixed number of environment transitions and we use this data to train the controller and the value function networks.
The value function network approximates the expected future returns of each state, and is defined for a policy $\pi$ as
\begin{align*}
V^\pi(o) = E_{o_0=o, a_t\sim \pi(\cdot | o_t)}\left[\sum_{t=0}^{\infty}\gamma^{\;t} r(o_t, a_t) \right].
\end{align*}
This function can be optimized using supervised learning due to its recursive nature:
\begin{align*}
    V^{\pi_{\theta}}(o_t) = \gamma\;V^{\pi_{\theta}}(o_{t+1}) + r_t,
\end{align*}
where
\begin{align*}
    V^{\pi_{\theta}}(o_T) = r_T + \gamma V^{\pi_{\theta_{old}}}(o_{T+1}).
\end{align*}
In PPO, the value function is used for computing the advantage
\begin{align*}
    A_t = V^{\pi_\theta} - V^{\pi_{\theta_{old}}}
\end{align*}
which is then used for training the policy by maximizing:
\begin{align*}
    L_{\pi}(\theta) = \frac{1}{T}\sum_{t=1}^T \min(\rho_t\hat{A}_t, \; \text{clip}(\rho_t,1-\epsilon,1+\epsilon)\hat{A}_t),
\end{align*}
where $\rho_t = \pi_{\theta}(a_t | o_t) \mathbin{/} \pi_{\theta_{old}}(a_t | o_t)$ is an importance sampling term used for calculating the expectation under the old policy $\pi_{\theta_{old}}$. 

\section{Training parameters}
\label{appendix:hyperparameters}
\begin{table}[H]
\centering
\small
\caption{Training details. We use the same parameters for every character.}
\begin{tabular}{lc}
\toprule
\bf{Hyperparameter} & \bf{Value} \\
\midrule
Learning Rate & $1.2 \times 10^{-4}$ \\
Optimizer & Adam \\
Batch Size & $8192$ \\
Num Environments & $4096$ \\
Episode Steps & $16$ \\
Num PPO Epochs & $5$ \\
Discount Factor $\gamma$ & $0.97$ \\
GAE-Lambda & $0.95$ \\
Value Loss Coefficient & $0.2$ \\
Clip parameter & $0.2$ \\
Max Grad Norm & $1.0$ \\ 
Exploration Noise & $0.02$ \\
Control time & 36 fps \\
Activation function & Tanh \\
Policy network & $[300, 200, 100]$ \\
Value network & $[400, 400, 300, 200]$ \\
\bottomrule
\end{tabular}
\end{table}

\section{Torque Limits}
\label{appendix:torque-limits}
\begin{table}[H]
\centering
\small
\caption{Torque limit scale value for each character's joint. If not written, then value is not scaled. Minimum value is negative of maximum value. Moreover, Oppy's tail and ears do not have torque values as they are passively actuated, similarly to final six joints of Dino's tail.}
\begin{tabular}{cccc}
\toprule
\bf{Parameter}           & \bf{Oppy} & \bf{Dino} & \bf{Jesse} \\
\midrule
Shoulder                 &  0.2   &  0.2      &  0.2         \\
Elbow                    &  0.1   &  0.1      &  0.1         \\
Head                     &  0.1   &  0.1      &  0.1         \\
Neck                     &  0.1   &  0.1      &  0.1         \\
Spine0                   &  1     &  1        &  0.25        \\
Spine1                   &  1     &  1        &  0.25        \\
Spine2                   &  1     &  1        &  0.25        \\
Spine3                   &  1     &  1        &  0.25        \\
Tail0                    &  -     &  0.5      &  -           \\
Tail1                    &  -     &  0.4      &  -           \\
\bottomrule
\end{tabular}
\end{table}

\end{document}